\documentclass[runningheads]{llncs}

\usepackage{eccv}

\usepackage{eccvabbrv}

\usepackage{graphicx}
\usepackage{booktabs}

\usepackage[accsupp]{axessibility}  

\usepackage{booktabs}
\usepackage{multirow}
\usepackage{xcolor}
\usepackage[table]{xcolor}
\usepackage{arydshln}
\usepackage{pifont}

\usepackage{wrapfig}
\graphicspath{{figures/}}

\usepackage{marvosym}

\newcommand{\cmark}{\ding{51}}
\newcommand{\xmark}{\ding{55}}
\newcommand{\equalcontrib}{\textsuperscript{\dag}}
\newcommand{\corremail}{\textsuperscript{\Letter}}
\usepackage{placeins}

\usepackage{hyperref}

\usepackage{orcidlink}

\begin{document}

\title{RA-SOD: Reliability-Aware RGB-T Salient Object Detection under Modality Degradation
} 

\titlerunning{RA-SOD}

\author{
Hongbo Gao\inst{1,2}\orcidlink{0009-0009-1826-1018}\equalcontrib \and
Zhengyu Li\inst{1}\orcidlink{0009-0005-9302-4718}\equalcontrib \and
Xueru Nie\inst{1}\orcidlink{0009-0004-2559-8333} \and
Dihao Zhu\inst{2}\orcidlink{0009-0002-3488-0643} \and
Lijun Zhao\inst{1}\orcidlink{0000-0002-9108-8276} \and
Yunke Wang\inst{2}\orcidlink{0009-0003-9796-530X}\corremail \and
Chang Xu\inst{2}\orcidlink{0000-0002-4756-0609}\corremail
}

\authorrunning{H. Gao et al.}

\institute{
State Key Laboratory of Robotics and Systems, Harbin Institute of Technology,
No.~92 West Dazhi Street, Nangang District, Harbin, Heilongjiang 150006, China
\and
School of Computer Science, University of Sydney, Camperdown, Sydney, NSW 2006, Australia\\
\email{\{23b908032,24s108484\}@stu.hit.edu.cn, 
\{yunke.wang,c.xu\}@sydney.edu.au}\\
\equalcontrib Equal contribution. \quad
\corremail Corresponding authors.
}

\maketitle

\begin{abstract}
RGB–Thermal (RGB-T) salient object detection leverages complementary cues from visible and thermal modalities to improve robustness in challenging environments. However, in real-world scenarios, the reliability of each modality is inherently unstable: RGB images degrade under low illumination, motion blur, and noise, while thermal imagery often suffers from contrast compression and sensor artifacts. Such degradation introduces unreliable perceptual evidence that can mislead cross-modal fusion and significantly deteriorate detection performance.
To address this challenge, we propose RA-SOD, a reliability-aware RGB-T salient object detection framework that explicitly models modality reliability and integrates it into feature learning and cross-modal fusion. First, we introduce a reliability-conditioned representation that adaptively compensates degraded modality features while preserving structural cues. Second, an uncertainty-guided dual-stream refinement strategy progressively corrects cross-modal representations while suppressing unreliable evidence. Finally, we propose a pixel-wise modality competition mechanism that dynamically selects modality cues according to spatial reliability for fine-grained fusion.
Extensive experiments on four benchmarks (VT821, VT1000, VT5000, and VT-IMAG) demonstrate that RA-SOD achieves state-of-the-art performance and exhibits strong robustness under severe modality degradation. Code and models are available at \url{https://github.com/zaoxienian/RA-SOD}.
\end{abstract}

\begin{figure}[tb]
  \centering
  \includegraphics[scale=0.35]{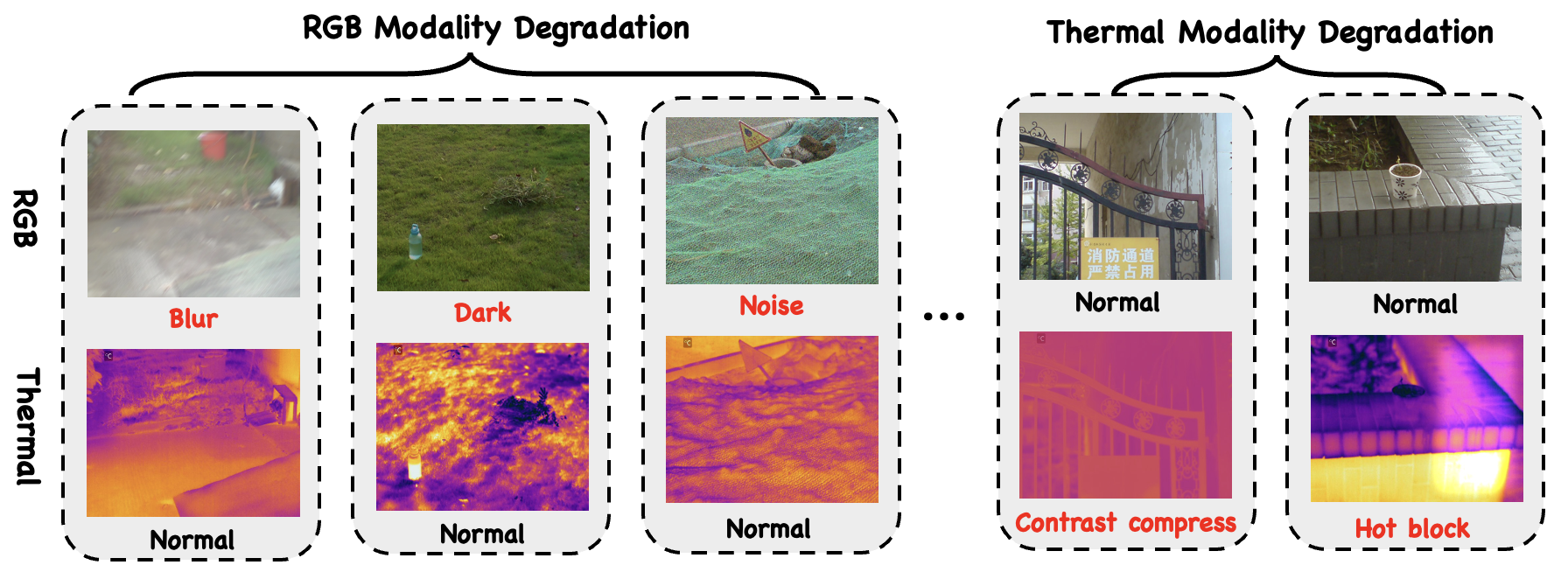}
  \caption{Examples of \textbf{modality degradation} in RGB–Thermal scenes. The RGB modality may suffer from blur, low illumination, and noise, while the thermal modality can exhibit contrast compression and block artifacts. Such degradations distort local responses and obscure object boundaries, posing significant challenges for reliable cross-modal fusion.}
  \label{fig:intro}
\end{figure}

\section{Introduction}
\label{sec:intro}

RGB–Thermal (RGB-T) salient object detection aims to identify visually prominent regions by jointly leveraging complementary information from visible and thermal infrared modalities \cite{wang2018rgb,tu2022rgbt}. Compared with single-modality approaches, RGB-T methods benefit from cross-modal complementarity, achieving improved robustness under low illumination and complex background conditions \cite{wang2023thermal,huo2021efficient,huo2022real,cong2022does}. However, in real-world environments, the quality of each modality is inherently unstable. The RGB modality may degrade under low-light conditions, motion blur, or heavy noise, while the thermal modality can suffer from contrast compression, suppressed temperature gradients, or block-like artifacts that distort local responses and obscure object boundaries. We refer to these phenomena as \textbf{modality degradation} \cite{hao2024cola,hu2025missingness,tian2025learning,wang2024rgb}, as shown in Fig. \ref{fig:intro}. Under such conditions, a modality may provide unreliable or even misleading perceptual evidence, which can adversely affect cross-modal fusion.

Most existing RGB-T salient object detection methods emphasize elaborate fusion strategies, such as two-stream architectures\cite{wang2025confidence,tang2024divide,8603756} or attention-based modality weighting mechanisms \cite{10032588,10504918}. These approaches implicitly assume that modality features extracted during encoding remain sufficiently reliable and cross-modal inconsistencies mainly arise at later fusion stages. However, in practice, modality degradation often emerges early and spreads through the network hierarchy \cite{han2022trusted,tu2021multi,tu2019m3s}. When degradation occurs, standard backbones apply fixed feature transformations that cannot adapt to input-dependent distributional shifts, causing noisy or distorted signals to be progressively propagated and even amplified. Consequently, subsequent fusion modules operate on already corrupted representations, making it difficult to disentangle structural information from modality-specific noise.

To address this issue, we propose a reliability-aware RGB-T salient object detection framework called RA-SOD. Instead of treating reliability as an auxiliary prior or a late-stage weighting cue, RA-SOD models reliability as a core principle and progressively propagates it across representation, refinement, and fusion stages.
\textbf{First}, at the encoding stage, we construct reliability-conditioned backbone representations through a parallel residual modulation branch with feature-conditioned routing. This design enables the network to adaptively compensate degraded modality features while preserving stable structural priors from the backbone.
\textbf{Second}, during decoding, we introduce an uncertainty-guided dual-stream recursive refinement scheme. Explicitly predicted uncertainty maps act as structured reliability signals to regulate bidirectional cross-modal correction, preventing unreliable features from being reinforced during hierarchical propagation.
\textbf{Finally}, at the fusion stage, we propose a hierarchical pixel-wise modality competition mechanism that dynamically selects RGB or thermal cues according to spatial reliability, enabling fine-grained cross-modal integration. By modeling and propagating reliability across representation, refinement, and fusion stages, RA-SOD establishes a unified hierarchical reliability framework. 

Extensive experiments on four RGB-T benchmarks (VT821, VT1000, VT5000, and VT-IMAG) demonstrate state-of-the-art performance and strong robustness under challenging modality degradation scenarios.

\section{Related Work}

\subsection{RGB Salient Object Detection}

Early RGB SOD methods rely on handcrafted features and heuristic priors (e.g., contrast~\cite{6871397,7307162} and center prior~\cite{6619110}), but show limited robustness in complex scenes. 
Deep learning, particularly fully convolutional networks (FCNs)~\cite{7488288}, later enables end-to-end pixel-wise prediction, leading to representative models such as Amulet~\cite{zhang2017amulet} and UCF~\cite{zhang2017learning}, with related efforts on weakly supervised object localization~\cite{mm/Xu000S023}. 
Subsequent works further enhance representation through multi-scale aggregation~\cite{8237695}, attention mechanisms~\cite{9076883,8954074}, and boundary guidance~\cite{8578428,zhao2019egnet,10006743}. 
More recently, generalist and state-space architectures have been explored, including VSCode~\cite{luo2024vscode} and Mamba-based Samba~\cite{11093604}. 
However, unimodal RGB methods remain vulnerable to adverse conditions such as low illumination, motion blur, and sensor noise.

\subsection{RGB-X Salient Object Detection}
Multi-modal SOD enhances robustness via auxiliary data. Specifically, RGB-D methods leverage geometric priors from depth maps to distinguish foreground from background \cite{7780626,8603756,wu2023hidanet}, optimizing detection results via multi-modal fusion networks (e.g., JL-DCF \cite{fu2020jl}, DFormer \cite{yin2025dformerv2}, \cite{10587282}). However, depth information is often compromised by low illumination or complex reflections.
Conversely, RGB-T SOD exploits thermal data to detect intrinsic radiation cues, benefiting from their robustness against lighting changes \cite{cong2022does,10127616}.
Existing works have transitioned from input-level concatenation to feature fusion mechanisms, involving feature interaction \cite{wang2024learning,wang2024alignment}, graph learning (e.g., MGAI \cite{10003255}, M3S-NIR \cite {tu2019m3s}, \cite{tu2020rgb}), and boundary enhancement(e.g., LSNet \cite{zhou2023lsnet}), which maximize cross-modal complementarity, and define the state-of-the-art for SOD.

\subsection{Multi-modal Learning in RGB-T SOD}
Existing RGB-T SOD methods typically formulate multi-modal learning as feature fusion~\cite{liu2023scribble,9161021,wan2024mffnet}, designing multi-scale modules to progressively integrate cross-modal information~\cite{gao2022unified,8935533,11131311}. 
Later works move beyond simple fusion to end-to-end representation and interaction modeling~\cite{10504918,10032588,wang2022cgfnet}. 
For example, Tu et al.~\cite{tu2021multi} propose a multi-interactive dual-decoder for cross-modal interaction, while ConTriNet~\cite{tang2024divide} introduces a divide-and-conquer triple-flow architecture, and SMR-Net~\cite{xiao2025smr} explores semantic-guided mutual reinforcement. 
Despite these advances, modality-degradation-aware representation learning remains less explored. Fixed feature transformations may propagate unreliable cues when one modality is degraded. Recent mixture-of-experts (MoE) models improve input-adaptive representation learning by routing features to specialized experts through learned gates~\cite{jacobs1991adaptive,shazeer2017outrageously,riquelme2021scaling}. 
Based on this input-adaptive routing mechanism, RA-SOD employs a lightweight shared expert bank to adapt RGB-T representations under modality degradation.

\begin{figure}[tb]
  \centering
  \includegraphics[scale=0.175]{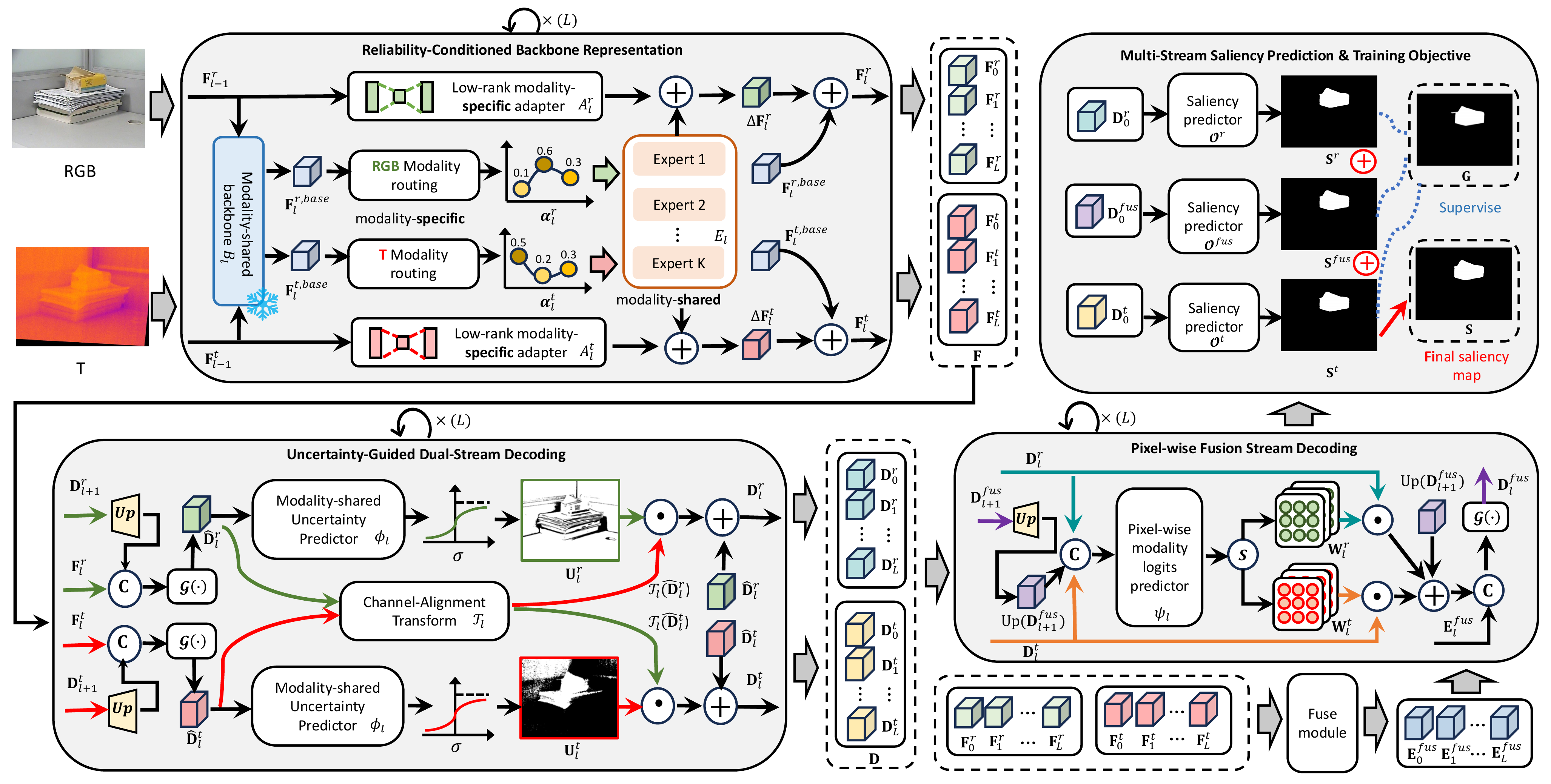}
  \caption{Overall architecture of RA-SOD. It consists of three key components: 
    (1) \textbf{Reliability-Conditioned Backbone Representation (RCBR)}: Extracts adaptive features by integrating structural priors from a frozen backbone with residual modulation via modality-specific adapters and a shared expert bank. 
    (2) \textbf{Uncertainty-Guided Dual-Stream Decoding}: Performs recursive cross-modal refinement regulated by predicted uncertainty maps to mitigate localized degradation. 
    (3) \textbf{Pixel-wise Fusion Stream Decoding}: Implements a competition mechanism to adaptively select modality-dependent cues at each spatial location. 
    The final saliency map is obtained by aggregating predictions from the RGB, Thermal, and Fusion streams.
  }
  \label{fig:method}
\end{figure}

\section{Methodology}
In this section, we introduce the framework of RA-SOD for RGB-T salient object detection under modality degradation.
The key idea is to model modality reliability and propagate it across representation, refinement, and fusion stages in the network. As illustrated in Fig.~\ref{fig:method},
RA-SOD consists of three main components:
(1) a \textbf{Reliability-Conditioned Backbone Representation (RCBR)}
that adaptively modulates modality features during encoding,
(2) an \textbf{Uncertainty-Guided Dual-Stream Decoding} mechanism
that progressively refines RGB and thermal representations
under reliability awareness, and
(3) a \textbf{Pixel-wise Fusion Stream Decoding} module
that performs spatially adaptive modality competition
for fine-grained cross-modal integration.
These components jointly form a hierarchical reliability modeling
framework that mitigates the propagation of unreliable cues
caused by modality degradation.

\subsection{Overall Architecture}
RA-SOD adopts a reliability-conditioned backbone
followed by a tri-stream hierarchical decoding architecture
for RGB-Thermal salient object detection.
Given aligned RGB and thermal inputs $I^r$ and $I^t$,
the network predicts a saliency map
$\mathbf{S}\in[0,1]^{H\times W}$.

Using a shared frozen backbone,
we first extract stage-wise base features
$\{\mathbf{F}^{r,\text{base}}_l\}_{l=0}^{L}$
and
$\{\mathbf{F}^{t,\text{base}}_l\}_{l=0}^{L}$.
Instead of directly using these features,
RA-SOD constructs reliability-conditioned representations
$\{\mathbf{F}^r_l\}_{l=0}^{L}$
and
$\{\mathbf{F}^t_l\}_{l=0}^{L}$
by introducing a parallel residual modulation branch
at each stage.
This design preserves a stable structural prior
from the frozen backbone
while enabling adaptive modulation under
modality-dependent reliability variations
(Sec.~\ref{sec3.2}).

Based on the resulting representations,
the network decodes three hierarchical streams in parallel:
an RGB stream, a Thermal stream, and a Fusion stream.
The RGB and Thermal streams perform uncertainty-guided
dual-stream decoding through recursive refinement,
while the Fusion stream conducts pixel-wise modality competition
within a top-down decoding hierarchy.
Each stream generates a saliency prediction
via a lightweight head with identical architecture
but independent parameters,
and the three predictions are linearly aggregated
to produce the final saliency map.

\subsection{Reliability-Conditioned Backbone Representation}
\label{sec3.2}

Reliable feature representation is critical under
modality degradation.
However, standard backbones apply fixed transformations
to all inputs, limiting their ability to adapt
to modality-dependent reliability variations.
To address this issue, we construct a
Reliability-Conditioned Backbone Representation (RCBR),
which augments the frozen backbone with a lightweight
residual modulation branch.
This design preserves stable structural priors
from the pretrained backbone while allowing
adaptive feature correction for degraded modalities.

Let $\mathbf{F}^m_{l-1}$ denote the input feature
of modality $m\in\{r,t\}$ at stage $l$.
The frozen backbone block produces a base feature

\begin{equation}
\mathbf{F}^{m,\text{base}}_l
=
B_l(\mathbf{F}^m_{l-1}),
\end{equation}

where $B_l(\cdot)$ denotes the shared frozen backbone transformation.

\subsubsection{Feature-conditioned modality routing.}

To adaptively regulate residual modulation,
we first predict expert routing weights
from the base feature:

\begin{equation}
\boldsymbol{\alpha}^m_l
=
\mathrm{Softmax}\!\left(
g^m_l(\mathbf{F}^{m,\text{base}}_l)
\right),
\qquad
\sum_{k=1}^{K}\alpha^m_{l,k}=1,
\label{eq:rcbr_gate_new}
\end{equation}

where $g^m_l(\cdot)$ is a lightweight gating function.
Since modality degradation often induces
distributional shifts in the backbone feature space,
conditioning routing decisions on
$\mathbf{F}^{m,\text{base}}_l$
allows the model to implicitly capture
modality-dependent reliability.

\subsubsection{MoE-guided residual representation adaptation.}

To enhance the expressive capacity of residual modulation,
we introduce a mixture-of-experts (MoE) mechanism
that provides multiple transformation subspaces.
Each expert captures a distinct feature adaptation pattern,
while routing weights dynamically determine
their contributions according to modality reliability.

In parallel to the frozen backbone transformation,
a residual branch operates directly on
$\mathbf{F}^m_{l-1}$:
\begin{equation}
\Delta \mathbf{F}^m_l
=
\sum_{k=1}^{K}
\alpha^m_{l,k}
E_{l,k}(\mathbf{F}^m_{l-1})
+
A^m_l(\mathbf{F}^m_{l-1}),
\label{eq:rcbr_res_new}
\end{equation}
where $E_{l,k}(\cdot)$ are shared expert transformations,
and $A^m_l(\cdot)$ is a modality-specific
low-rank adapter.

The final reliability-conditioned representation is
\begin{equation}
\mathbf{F}^m_l
=
\mathbf{F}^{m,\text{base}}_l
+
\Delta \mathbf{F}^m_l.
\label{eq:rcbr_final}
\end{equation}
This parallel design preserves the stable structural prior
of the frozen backbone while enabling adaptive
modulation of unreliable feature components.
In the subsequent decoding stage (Sec.~3.3),
reliability is further modeled explicitly
via uncertainty-guided cross-modal refinement.

\subsection{Uncertainty-Guided Dual-Stream Decoding}

During decoding, modality degradation may lead to
spatially localized ambiguity,
where certain regions contain unreliable cues.
Directly propagating such features can amplify errors
during hierarchical decoding.
To alleviate this issue, we introduce an
Uncertainty-Guided Dual-Stream Decoding mechanism,
which explicitly estimates uncertainty maps
to regulate cross-modal feature correction.

Specifically, RGB and Thermal modalities are decoded
through two parallel top-down streams.
At stage $l$, a preliminary decoding feature
$\hat{\mathbf{D}}^m_l$ is computed from
$\mathbf{F}^m_l$,
upon which bidirectional uncertainty-guided refinement is applied.
The refined state $\mathbf{D}^m_l$
is propagated to the next finer stage,
forming a hierarchical recursion.

\subsubsection{Top-down recursion.}

Given the reliability-conditioned encoder representations,
we first construct preliminary decoding features
through a top-down hierarchical decoding process.

At the coarsest stage $L$,

\begin{equation}
\hat{\mathbf{D}}^r_L = \mathcal{G}^r_L(\mathbf{F}^r_L),
\quad
\hat{\mathbf{D}}^t_L = \mathcal{G}^t_L(\mathbf{F}^t_L).
\end{equation}

For $l=L-1,\dots,0$,

\begin{equation}
\hat{\mathbf{D}}^m_l
=
\mathcal{G}^m_l\!\left(
[\mathbf{F}^m_l;\mathrm{Up}(\mathbf{D}^m_{l+1})]
\right),
\quad m\in\{r,t\},
\label{eq:dual_prelim}
\end{equation}

where $[\cdot;\cdot]$ denotes channel concatenation
and $\mathrm{Up}(\cdot)$ represents bilinear upsampling.

\subsubsection{Uncertainty-guided cross-modal refinement.}

Although the preliminary decoding features provide
modality-specific predictions,
their reliability may vary across spatial locations
under modality degradation.
To explicitly model such reliability variations,
we introduce an uncertainty-guided refinement mechanism.

An uncertainty map is predicted as
\begin{equation}
\mathbf{U}^m_l
=
\sigma\!\left(
\phi_l(\hat{\mathbf{D}}^m_l)
\right),
\label{eq:dual_unc}
\end{equation}
where $\phi_l(\cdot)$ is a modality-shared lightweight uncertainty predictor.

The uncertainty map regulates cross-modal correction
\begin{equation}
\mathbf{D}^r_l
=
\hat{\mathbf{D}}^r_l
+
\mathbf{U}^r_l \odot \mathcal{T}_l(\hat{\mathbf{D}}^t_l),
\quad
\mathbf{D}^t_l
=
\hat{\mathbf{D}}^t_l
+
\mathbf{U}^t_l \odot \mathcal{T}_l(\hat{\mathbf{D}}^r_l),
\label{eq:dual_refine}
\end{equation}
where $\mathcal{T}_l(\cdot)$ denotes a channel-alignment transform.

The refined states are recursively propagated
to form decoding pyramids
$\{\mathbf{D}^r_l\}$ and $\{\mathbf{D}^t_l\}$.

\subsection{Pixel-wise Fusion Stream Decoding}

Besides modality-specific decoding pyramids,
a third fusion stream
$\{\mathbf{D}^{fus}_l\}_{l=0}^{L}$
is decoded hierarchically.
Encoder features are first fused through the
MFM module~\cite{tang2024divide},
producing modality-integrated encoder representations
$\{\mathbf{E}^{fus}_l\}_{l=0}^{L}$,
which serve as the input to the fusion decoding process.

\subsubsection{Fusion recursion.}

Starting from the coarsest level,
the fusion stream progressively aggregates
encoder features and modality-guided cues
in a top-down manner:
\begin{equation}
\mathbf{D}^{fus}_l
=
\mathcal{G}^{fus}_l
\!\left(
[\mathbf{E}^{fus}_l;\mathbf{H}_l]
\right),
\quad l=L,\dots,0,
\end{equation}
with $\mathbf{H}_L=\mathbf{0}$.

\subsubsection{Pixel-wise modality competition.}

Although the dual-stream decoding stage produces refined
RGB and thermal representations,
their reliability may still vary across spatial locations
under modality degradation.
Therefore, instead of using fixed fusion weights,
we introduce a pixel-wise modality competition mechanism
to dynamically determine which modality provides
more reliable cues at each location.

For $l=L-1,\dots,0$,
\begin{equation}
[\mathbf{W}^r_l,\mathbf{W}^t_l]
=
\mathrm{Softmax}
\!\left(
\psi_l(
[\mathrm{Up}(\mathbf{D}^{fus}_{l+1});
\mathbf{D}^r_l;
\mathbf{D}^t_l]
)
\right),
\end{equation}

where $\psi_l(\cdot)$ predicts modality logits
at each spatial location. 

The fusion guidance feature is
\begin{equation}
\mathbf{H}_l
=
\mathbf{W}^r_l \odot \mathbf{D}^r_l
+
\mathbf{W}^t_l \odot \mathbf{D}^t_l
+
\mathrm{Up}(\mathbf{D}^{fus}_{l+1}).
\end{equation}

This design allows the model to adaptively emphasize
the more reliable modality while suppressing unreliable
responses caused by local degradation.

\subsection{Multi-Stream Saliency Prediction}

After hierarchical decoding, each stream produces
a modality-specific representation that captures
complementary saliency cues.
Therefore, we attach an independent prediction head
to each stream to generate saliency maps.
Each stream predicts saliency independently:
\begin{equation}
\mathbf{S}^r
=
\mathcal{O}^r(\mathbf{D}^r_0),
\quad
\mathbf{S}^t
=
\mathcal{O}^t(\mathbf{D}^t_0),
\quad
\mathbf{S}^{fus}
=
\mathcal{O}^{fus}(\mathbf{D}^{fus}_0),
\end{equation}
where $\mathcal{O}^r$, $\mathcal{O}^t$, and $\mathcal{O}^{fus}$
denote lightweight prediction heads.
They share the same architecture but have independent parameters.
The final saliency map is obtained via linear aggregation:
\begin{equation}
\mathbf{S}
=
\mathbf{S}^r
+
\mathbf{S}^t
+
\mathbf{S}^{fus}.
\end{equation}
This multi-stream prediction strategy enables the model
to exploit complementary cues from modality-specific
and fused representations.

\subsection{Training Objective}

To encourage reliability-aware learning
while avoiding modality dominance,
we impose deep supervision on
the RGB, Thermal, and Fusion predictions.
The three streams are optimized jointly,
and the overall objective is
\begin{equation}
\mathcal{L}
=
\sum_{m\in\{r,t,fus\}}
\left(
\lambda_1
\mathcal{L}_{\mathrm{BCE}}(\mathbf{S}^m,\mathbf{G})
+
\lambda_2
\mathcal{L}_{\mathrm{IoU}}(\mathbf{S}^m,\mathbf{G})
\right),
\end{equation}
where $\mathbf{G}$ is the ground-truth map.
This multi-stream supervision stabilizes optimization
and yields complementary saliency cues
that improve the final aggregated prediction.

\section{ Experiments}

\subsection{Experimental Setup}

\subsubsection{Implementation Details.}

Experiments are conducted on a single NVIDIA RTX 3090 GPU.
Following ConTriNet~\cite{tang2024divide}, we adopt an ImageNet-pretrained Res2Net-50 as the shared backbone.
RGB and thermal images are resized to $352\times352$ with random flipping, rotation, and border cropping for augmentation.
The network is optimized with Adam using an initial learning rate of $5\times10^{-5}$ and a cosine decay schedule for 100 epochs.
During inference, saliency maps are resized to the original resolution and the final prediction is obtained by linearly aggregating the three outputs.
Performance is evaluated using the Saliency-Evaluation-Toolbox.

\subsubsection{Datasets.}
We evaluate our method on three RGB-T benchmarks: VT821~\cite{wang2018rgb}, VT1000~\cite{tu2020rgb}, and VT5000~\cite{tu2022rgbt}. 
To further assess robustness under severe degradations, we additionally adopt the challenging VT-IMAG~\cite{tang2024divide}. 
Following ConTriNet~\cite{tang2024divide}, 2,500 pairs from VT5000 are used for training, and the remaining 2,500 pairs together with the full VT821, VT1000, and VT-IMAG datasets are used for evaluation.

\begin{table*}[t]
  \centering
  \caption{Comparison with recent state-of-the-art CNN-based methods in RGB-D/RGB-T SOD on VT821, VT1000 and VT5000 benchmarks. The best and second-best results are highlighted in \textcolor{red}{red} and \textcolor{blue}{blue}, respectively.}
  \label{tab:sota_comparison}
  \resizebox{\textwidth}{!}{
  \begin{tabular}{l ccccc ccccc ccccc}
    \toprule
    \multirow{2}{*}{Method} & \multicolumn{5}{c}{VT821} & \multicolumn{5}{c}{VT1000} & \multicolumn{5}{c}{VT5000} \\
    \cmidrule(lr){2-6} \cmidrule(lr){7-11} \cmidrule(lr){12-16}
    & $S_m\uparrow$ & $F_\beta\uparrow$ & $F_\beta^w\uparrow$ & $E_m\uparrow$ & $\mathcal{M}\downarrow$ & $S_m\uparrow$ & $F_\beta\uparrow$ & $F_\beta^w\uparrow$ & $E_m\uparrow$ & $\mathcal{M}\downarrow$ & $S_m\uparrow$ & $F_\beta\uparrow$ & $F_\beta^w\uparrow$ & $E_m\uparrow$ & $\mathcal{M}\downarrow$ \\
    \midrule
    DCMF~\cite{wang2022learning} & 0.856 & 0.834 & 0.740 & 0.866 & 0.055 & 0.917 & 0.834 & 0.841 & 0.915 & 0.028 & 0.857 & 0.753 & 0.728 & 0.873 & 0.052 \\
    CIR-Net~\cite{cong2022cir} & 0.861 & 0.777 & 0.724 & 0.884 & 0.045 & 0.914 & 0.861 & 0.823 & 0.924 & 0.028 & 0.871 & 0.791 & 0.744 & 0.899 & 0.041 \\
    RAFNet~\cite{wu2022robust} & 0.883 & 0.826 & 0.811 & 0.905 & 0.038 & \textcolor{blue}{0.932} & 0.893 & 0.898 & 0.941 & 0.020 & 0.891 & 0.840 & 0.829 & 0.921 & 0.033 \\
    PopNet~\cite{wu2023source} & 0.861 & 0.795 & 0.774 & 0.887 & 0.043 & 0.929 & 0.893 & 0.895 & 0.939 & 0.020 & 0.887 & 0.839 & 0.827 & 0.918 & 0.034 \\
    HiDAnet~\cite{wu2023hidanet} & 0.890 & 0.842 & 0.832 & 0.920 & \textcolor{blue}{0.031} & 0.927 & 0.894 & 0.897 & 0.942 & 0.020 & 0.890 & 0.846 & 0.836 & 0.923 & 0.032 \\
    \midrule
    MIA~\cite{liang2022multi} & 0.844 & 0.740 & 0.720 & 0.850 & 0.070 & 0.924 & 0.868 & 0.864 & 0.926 & 0.025 & 0.878 & 0.793 & 0.780 & 0.893 & 0.040 \\
    ECFFNet~\cite{zhou2022ecffnet} & 0.877 & 0.810 & 0.801 & 0.902 & 0.034 & 0.923 & 0.876 & 0.885 & 0.930 & 0.021 & 0.874 & 0.806 & 0.801 & 0.906 & 0.038 \\
    OSRNet~\cite{huo2022real} & 0.875 & 0.813 & 0.801 & 0.896 & 0.043 & 0.926 & 0.892 & 0.891 & 0.935 & 0.022 & 0.875 & 0.823 & 0.807 & 0.908 & 0.040 \\
    LSNet~\cite{zhou2023lsnet} & 0.878 & 0.825 & 0.809 & 0.911 & 0.033 & 0.925 & 0.885 & 0.887 & 0.935 & 0.023 & 0.877 & 0.825 & 0.806 & 0.915 & 0.037 \\
    CAVER~\cite{pang2023caver} & \textcolor{blue}{0.891} & 0.839 & 0.835 & 0.919 & 0.033 & \textcolor{red}{0.936} & 0.903 & \textcolor{blue}{0.909} & 0.945 & \textcolor{red}{0.017} & 0.892 & 0.842 & 0.835 & 0.924 & 0.032 \\
    LAFB~\cite{wang2024learning} & 0.887 & 0.843 & 0.817 & 0.915 & \textcolor{red}{0.027} &0.927  & 0.905 & 0.905 & 0.945 & \textcolor{blue}{0.018} & 0.890 & 0.857 & 0.841 & 0.931 & \textcolor{blue}{0.030} \\
    SMR-Net~\cite{xiao2025smr} & 0.889 &\textcolor{blue}{0.846} & 0.835 & \textcolor{blue}{0.921} & \textcolor{blue}{0.031} & 0.928 & 0.902 & 0.908 & 0.947 & 0.019 & 0.893 & 0.859 & 0.845 & \textcolor{blue}{0.935} & \textcolor{blue}{0.030} \\
    
    ConTriNet~\cite{tang2024divide} & 0.883 & \textcolor{blue}{0.846} & \textcolor{blue}{0.836} & 0.920 & 0.033 & 0.927 & \textcolor{blue}{0.903} & 0.903 & \textcolor{blue}{0.948} & 0.019 & \textcolor{blue}{0.894} & \textcolor{blue}{0.860} & \textcolor{blue}{0.846} & 0.934 & \textcolor{blue}{0.030} \\
    \midrule
    \textbf{RA-SOD (Ours)} & \textcolor{red}{0.892} & \textcolor{red}{0.857} & \textcolor{red}{0.846} & \textcolor{red}{0.925} & \textcolor{blue}{0.031} & \textcolor{blue}{0.932} & \textcolor{red}{0.915} & \textcolor{red}{0.913} & \textcolor{red}{0.951} & \textcolor{blue}{0.018} & \textcolor{red}{0.897} & \textcolor{red}{0.870} & \textcolor{red}{0.855} & \textcolor{red}{0.937} & \textcolor{red}{0.029} \\
    \bottomrule
  \end{tabular}
  }
\end{table*}

\subsubsection{Evaluation Metrics.}

Specifically, we adopt five standard metrics: Structure-measure ($S_{m}$)~\cite{fan2017structure}, Mean F-measure ($F_{\beta}$)~\cite{achanta2009frequency}, Weighted F-measure ($F_{\beta}^{w}$)~\cite{margolin2014how}, Mean Enhanced-alignment measure ($E_{m}$)~\cite{fan2018enhanced}, and Mean Absolute Error ($\mathcal{M}$)~\cite{perazzi2012saliency}. 
Among them, $S_{m}$ evaluates structural similarity; $F_{\beta}$ and $F_{\beta}^{w}$ measure the precision–recall trade-off; $E_{m}$ considers both pixel-level alignment and global statistics; and $\mathcal{M}$ computes the average pixel-wise error. 
We report mean $F_{\beta}$ and $E_{m}$ scores averaged over all thresholds. 
Higher values indicate better performance for all metrics except $\mathcal{M}$.

\subsection{Comparison on Standard Benchmarks}
We compare RA-SOD with several recent state-of-the-art methods,
including five RGB-D SOD models
(DCMF~\cite{wang2022learning}, CIR-Net~\cite{cong2022cir}, RAFNet~\cite{wu2022robust},
PopNet~\cite{wu2023source}, and HiDAnet~\cite{wu2023hidanet})
and eight RGB-T SOD models
(MIA~\cite{liang2022multi}, ECFFNet~\cite{zhou2022ecffnet}, OSRNet~\cite{huo2022real},
LSNet~\cite{zhou2023lsnet}, CAVER~\cite{pang2023caver}, LAFB~\cite{wang2024learning},
SMR-Net~\cite{xiao2025smr}, and ConTriNet~\cite{tang2024divide}).
For fair comparison, all RGB-T methods are retrained on the RGB-T dataset using their official implementations and consistent experimental settings, while RGB-D models follow their default hyperparameters.
For SMR-Net~\cite{xiao2025smr}, we replace the original ResNet-30 backbone with Res2Net-50~\cite{he2019bag} to ensure a fair evaluation.

\subsubsection{Quantitative Results.}
Table~\ref{tab:sota_comparison} reports quantitative comparisons under five evaluation metrics. The results show that RA-SOD achieves the best or second-best performance for all five metrics on all three benchmarks.
On the large-scale VT5000 benchmark, RA-SOD ranks first on every metric. Compared with ConTriNet~\cite{tang2024divide}, RA-SOD improves $F_{\beta}$ by 0.010 and $F_{\beta}^{w}$ by 0.009, and also brings a 0.003 gain in $E_{m}$, while reducing $\mathcal{M}$ by 0.001.
On the smaller VT821 and VT1000 benchmarks, RA-SOD remains consistently within the top two. 
On VT1000, RA-SOD achieves the best $F_{\beta}$, $F_{\beta}^{w}$, and $E_{m}$, and it stays highly competitive in structure measure and $\mathcal{M}$ compared with CAVER~\cite{pang2023caver}.
Overall, these results demonstrate strong generalization from small and medium-scale benchmarks to the larger and more diverse VT5000.

\begin{figure}[tb]
  \centering
  \includegraphics[scale=0.38]{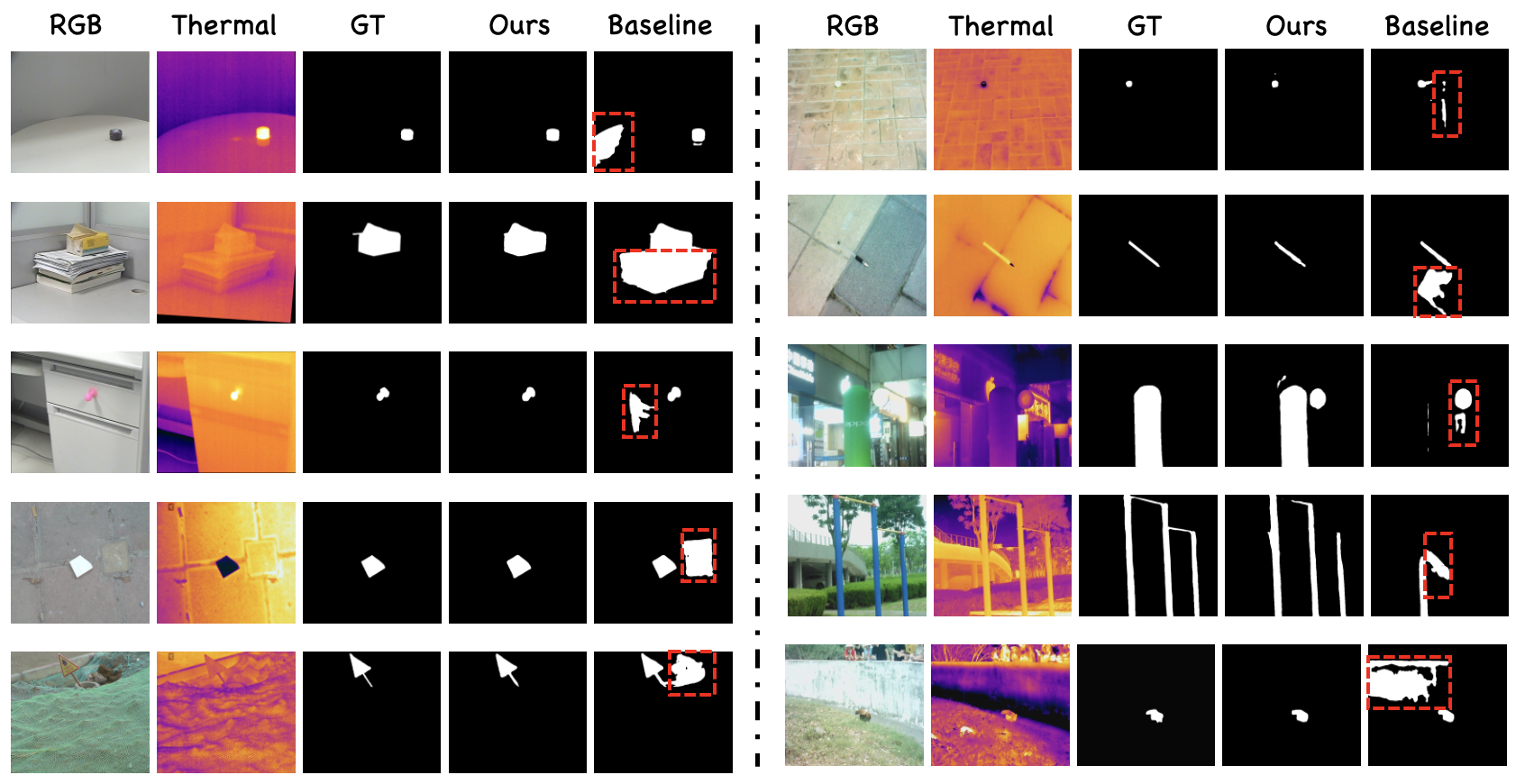}
  \caption{Qualitative comparison between RA-SOD (ours) and ConTriNet (baseline) on standard RGB-T SOD benchmarks (VT821, VT1000, and VT5000). Red dashed boxes highlight typical failure cases of the baseline.}
  \vskip -0.15in
  \label{fig:visual1}
\end{figure}

\begin{figure}[tb]
  \centering
  \includegraphics[scale=0.45]{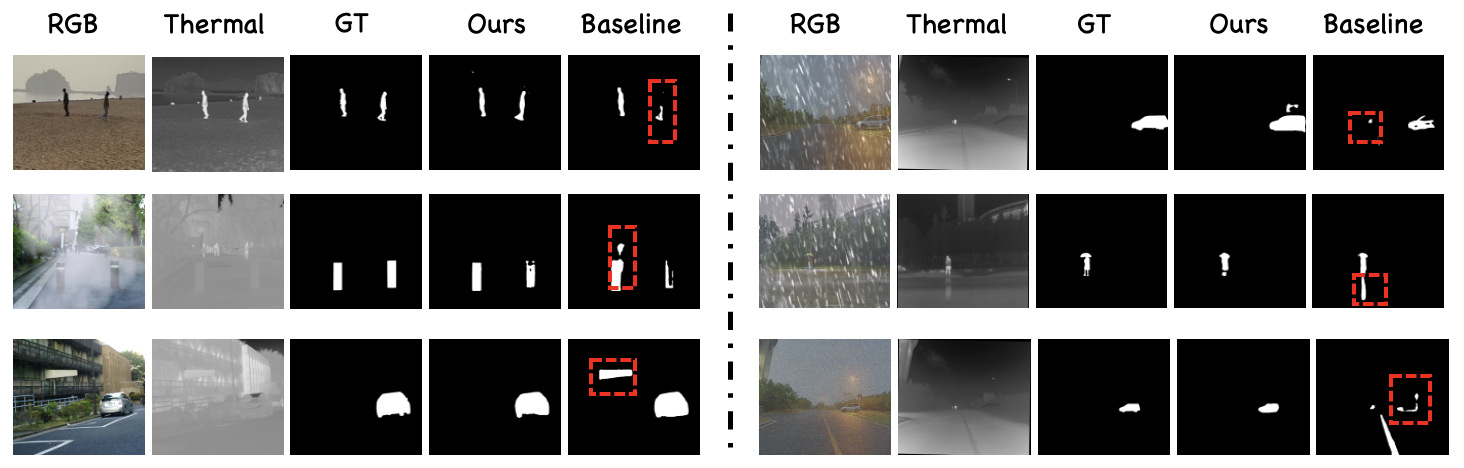}
  \caption{Qualitative comparison between RA-SOD (ours) and ConTriNet (baseline) on the challenging VT-IMAG dataset. 
}
  \vskip -0.15in
  \label{fig:vtim}
\end{figure}

\subsubsection{Qualitative Results and Visualization.}

The visual comparisons in Fig.~\ref{fig:visual1} reveal the limitations of ConTriNet~\cite{tang2024divide} (Baseline).
Without explicit reliability modeling, it struggles to distinguish salient targets from background noise, leading to severe false detections (red dashed boxes).
In contrast, our reliability-aware framework effectively suppresses unreliable cues, enabling accurate localization with complete structures and sharp boundaries even under extremely low signal-to-noise conditions.

\subsection{Robustness Analysis under Degradations}

\subsubsection{Synthetic Degradation Settings.}
To verify the effectiveness of reliability awareness under modality degradation, we construct synthetic degraded test sets derived from the VT1000 dataset~\cite{tu2020rgb} by degrading only one modality at a time while keeping the other modality and the ground-truth masks unchanged. We focus on five representative degradations, including low-light, additive noise, and blur on the RGB modality, as well as local hot-block artifacts and global contrast compression on the thermal modality. The degradation severity is randomly sampled per image from preset ranges with a fixed seed to ensure reproducibility.

\subsubsection{Comparison under Synthetic Degradations.}
\begin{wraptable}[8]{r}{0.52\textwidth}
  \vskip -0.42in
  \centering
  \caption{Average comparison under synthetic degradations on VT1000.}
  \label{tab:synth_deg_vt1000}
  \scriptsize
  \setlength{\tabcolsep}{1.8pt}
  \begin{tabular}{lccccc}
    \toprule
    Method & $S_m\uparrow$ & $F_{\beta}\uparrow$ & $F_{\beta}^{w}\uparrow$ & $E_m\uparrow$ & $\mathcal{M}\downarrow$ \\
    \midrule
    LAFB~\cite{wang2024learning} & 0.919 & 0.887 & 0.887 & 0.938 & 0.022 \\
    SMR-Net~\cite{xiao2025smr} & 0.917 & 0.895 & 0.893 & 0.945 & 0.021 \\
    ConTriNet~\cite{tang2024divide} & 0.921 & 0.893 & 0.895 & 0.945 & 0.021 \\
    \midrule
    \cellcolor[gray]{0.9}\textbf{RA-SOD} & \cellcolor[gray]{0.9}\textbf{0.927} & \cellcolor[gray]{0.9}\textbf{0.906} & \cellcolor[gray]{0.9}\textbf{0.907} & \cellcolor[gray]{0.9}\textbf{0.949} & \cellcolor[gray]{0.9}\textbf{0.020} \\
    \bottomrule
  \end{tabular}
  \vskip -0.05in
\end{wraptable}

Table~\ref{tab:synth_deg_vt1000} compares RA-SOD with representative RGB-T SOD methods under synthetic degradations. Existing methods suffer noticeable performance drops in corrupted scenarios due to limited mechanisms for handling unreliable modality features. In contrast, RA-SOD consistently improves performance by explicitly modeling modality reliability. Compared with the strongest competing results, RA-SOD improves $S_m$ from 0.921 to 0.927, $F_{\beta}$ from 0.895 to 0.906, and $F_{\beta}^{w}$ from 0.895 to 0.907, while reducing $\mathcal{M}$ to 0.020. These results demonstrate the effectiveness of reliability-aware modeling for robust RGB-T saliency detection under modality degradation.

\subsubsection{Performance on the VT-IMAG Challenge Set.}

Table~\ref{tab:vtimag_comparison_cnn} reports the quantitative comparison on the challenging VT-IMAG benchmark~\cite{tang2024divide}, which is designed to evaluate robustness under severe sensor failures and environmental degradations (e.g., thermal crossover, bad weather, and strong noise). 
RA-SOD achieves superior performance on four out of five metrics among state-of-the-art CNN-based methods. 
Notably, it significantly outperforms the strong baseline ConTriNet~\cite{tang2024divide} on its own benchmark, improving $F_{\beta}$ by $0.020$ and $F_{\beta}^{w}$ by $0.008$. 
These results indicate that while the baseline struggles to preserve structural integrity under heavy modality corruption, RA-SOD effectively suppresses unreliable cues via reliability-aware modeling, achieving the highest $S_m$ of $0.829$ and $E_m$ of $0.908$. 
The qualitative comparisons in Fig.~\ref{fig:vtim} further demonstrate that RA-SOD produces more reliable and complete saliency predictions under severe degradation.

\begin{table}[t]
  \centering
  \caption{Quantitative comparison with state-of-the-art methods on the VT-IMAG benchmark. \textcolor{red}{\text{Red}} and \textcolor{blue}{\text{Blue}} indicate the best and second-best performance, respectively.}
  \label{tab:vtimag_comparison_cnn}
  \setlength{\tabcolsep}{6pt}
  \small
  \begin{tabular}{l c c c c c}
    \toprule
    Method & $S_m \uparrow$ & $F_{\beta} \uparrow$ & $F_{\beta}^{w} \uparrow$ & $E_m \uparrow$ & $\mathcal{M} \downarrow$ \\
    \midrule
    CGFNet~\cite{wang2022cgfnet} & 0.795 & 0.696 & 0.664 & 0.861 & 0.042 \\
    OSRNet~\cite{huo2022real}    & 0.782 & 0.665 & 0.626 & 0.839 & 0.051 \\
    LSNet~\cite{zhou2023lsnet}   & 0.779 & 0.712 & 0.636 & 0.896 & 0.038 \\

    CAVER~\cite{pang2023caver}   & 0.815 & 0.724 & 0.693 & 0.885 & 0.032 \\
    ConTriNet~\cite{tang2024divide} & \textcolor{blue}{\text{0.828}} & \textcolor{blue}{\text{0.745}} & \textcolor{blue}{\text{0.727}} & \textcolor{blue}{\text{0.902}} & \textcolor{red}{\text{0.029}} \\
    \midrule
    \cellcolor[gray]{0.9}\textbf{RA-SOD (Ours)} & \cellcolor[gray]{0.9}\textcolor{red}{\text{0.829}} & \cellcolor[gray]{0.9}\textcolor{red}{\text{0.765}} & \cellcolor[gray]{0.9}\textcolor{red}{\text{0.735}} & \cellcolor[gray]{0.9}\textcolor{red}{\text{0.908}} & \cellcolor[gray]{0.9}\textcolor{blue}{\text{0.031}} \\
    \bottomrule
  \end{tabular}
\end{table}

\subsection{Analysis of RA-SOD}

\subsubsection{Model Complexity and Efficiency.}
\begin{wraptable}[8]{r}{0.52\textwidth}
  \vskip -0.45in
  \centering
  \caption{Comparison of model complexity and inference speed.}
  \label{tab:efficiency}
  \scriptsize
  \setlength{\tabcolsep}{1.8pt}
  \begin{tabular}{lcccc}
    \toprule
    Method & Backbone & Params. & FLOPs & FPS \\
    \midrule
    ConTriNet~\cite{tang2024divide} & R2N-50 & 34.78 & 55.42 & 13.53 \\
    LAFB~\cite{wang2024learning} & R2N-50 & 453.03 & 139.70 & 4.62 \\
    SMR-Net~\cite{xiao2025smr} & R2N-50 & 47.70 & 62.80 & 10.15 \\
    \midrule
    \rowcolor[gray]{0.9}
    \textbf{RA-SOD} & R2N-50 & \textbf{46.72} & \textbf{60.13} & \textbf{11.24} \\
    \bottomrule
  \end{tabular}
  \vskip -0.08in
\end{wraptable}

Table~\ref{tab:efficiency} compares the model complexity of different methods. For fair evaluation, all compared methods are implemented with a Res2Net-50 backbone. ConTriNet~\cite{tang2024divide} is the most lightweight model with 34.78M parameters and 55.42G FLOPs, while RA-SOD ranks second with 46.72M parameters and 60.13G FLOPs. RA-SOD also runs faster than SMR-Net and LAFB, showing a good balance between complexity and performance.

\subsubsection{Backbone Generalization.}
Table~\ref{tab:backbone_transformer} reports the comparison with Transformer-based RGB-T SOD methods on VT5000. With Swin-B, RA-SOD$^\dagger$ achieves the best results on all five metrics and clearly outperforms ConTriNet. These results show that the proposed reliability-aware design can generalize from Res2Net-50 to Transformer-based backbones.

\begin{table}[!t]
  \centering
  \caption{Transformer-backbone comparison on VT5000. RA-SOD$^\dagger$ uses Swin-B. ``--'' denotes metrics not reported.}
  \label{tab:backbone_transformer}
  \small
  \setlength{\tabcolsep}{3.5pt}
  \begin{tabular}{l l c c c c c}
    \toprule
    Method & Backbone & $S_m\uparrow$ & $F_{\beta}\uparrow$ & $F_{\beta}^{w}\uparrow$ & $E_m\uparrow$ & $\mathcal{M}\downarrow$ \\
    \midrule
    HRTransNet\cite{tang2022hrtransnet} & HRFormer & 0.912 & 0.871 & 0.870 & 0.945 & 0.025 \\
    XMSNet\cite{wu2023object} & Transformer & 0.907 & 0.871 & 0.865 & 0.939 & 0.028 \\
    SACNet\cite{wang2024alignment} & Swin-B & 0.917 & -- & 0.888 & 0.957 & 0.021 \\
    PCNet\cite{wang2025alignment} & Swin-B & 0.920 & 0.899 & -- & 0.956 & -- \\
    ConTriNet\cite{tang2024divide} & Swin-B & 0.923 & 0.898 & 0.895 & 0.956 & 0.020 \\
    \rowcolor[gray]{0.9}
    \textbf{RA-SOD$^\dagger$} & Swin-B & \textbf{0.930} & \textbf{0.904} & \textbf{0.899} & \textbf{0.963} & \textbf{0.017} \\
    \bottomrule
  \end{tabular}
  \vskip -0.05in
\end{table}

\begin{figure}[tb]
  \centering
  \includegraphics[scale=0.37]{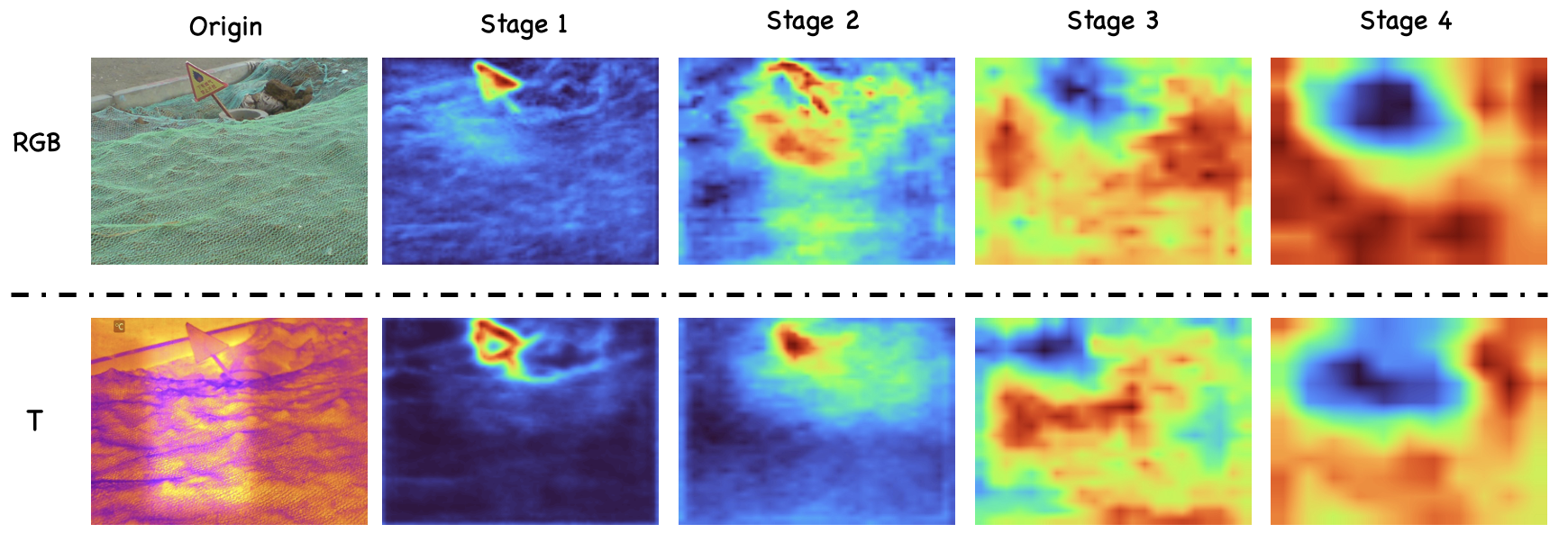}
  \caption{Visualization of the predicted uncertainty maps across four decoding stages for the RGB (top) and Thermal (bottom) branches. Brighter regions indicate higher predicted uncertainty. The evolution of these maps demonstrates the model's progressive refinement from boundary-level ambiguity to region-level reliability.
  }
  \label{fig:uncertain}
\end{figure}

\subsubsection{Uncertainty Visualization.}
Fig.~\ref{fig:uncertain} visualizes the predicted uncertainty maps at different decoding stages, where brighter regions indicate higher uncertainty. 
At early stages, uncertainty mainly appears around object boundaries due to limited semantic capacity and foreground–background ambiguity. 
As decoding progresses, uncertainty within object interiors gradually decreases, reflecting more stable object-level predictions from deeper features. 
Meanwhile, RGB and thermal streams exhibit distinct uncertainty patterns, suggesting that the learned uncertainty captures modality-dependent reliability rather than random noise.

\subsection{Ablation Studies}

\begin{table*}[t]
  \centering
  \caption{Ablation study of the Reliability-Conditioned Backbone Representation (RCBR) on three benchmarks. ``A'' denotes the low-rank modality-specific adapter, and ``E'' represents the shared expert bank (MoE). B0 is the base network without residual modulation. The best results are highlighted in \textbf{bold}.}
  \label{tab:abrcbr}
  \resizebox{\textwidth}{!}{
  \begin{tabular}{l cc ccccc ccccc ccccc}
    \toprule
    \multirow{2}{*}{Variant} & \multicolumn{2}{c}{Module} & \multicolumn{5}{c}{VT821} & \multicolumn{5}{c}{VT1000} & \multicolumn{5}{c}{VT5000} \\
    \cmidrule(lr){2-3} \cmidrule(lr){4-8} \cmidrule(lr){9-13} \cmidrule(lr){14-18}
    & A & E & $S_m\uparrow$ & $F_\beta\uparrow$ & $F_\beta^w\uparrow$ & $E_m\uparrow$ & $M\downarrow$ & $S_m\uparrow$ & $F_\beta\uparrow$ & $F_\beta^w\uparrow$ & $E_m\uparrow$ & $M\downarrow$ & $S_m\uparrow$ & $F_\beta\uparrow$ & $F_\beta^w\uparrow$ & $E_m\uparrow$ & $M\downarrow$ \\
    \midrule
    Frozen Backbone (B0) & $\times$ & $\times$ & 0.886 & 0.852 & 0.840 & 0.924 & 0.032 & 0.929 & 0.906 & 0.906 & 0.948 & 0.019 & 0.896 & 0.865 & 0.851 & 0.935 & 0.030 \\
    +A only (B1)         & \checkmark & $\times$ & 0.888 & 0.856 & 0.843 & \textbf{0.925} & 0.032 & \textbf{0.933} & \textbf{0.916} & \textbf{0.914} & \textbf{0.951} & \textbf{0.018} & 0.896 & 0.868 & 0.853 & 0.936 & 0.031 \\
    +E only (B2)         & $\times$ & \checkmark & 0.888 & \textbf{0.858} & 0.842 & 0.924 & 0.032 & 0.931 & 0.910 & 0.910 & 0.949 & 0.020 & 0.896 & 0.866 & 0.851 & \textbf{0.937} & \textbf{0.029} \\
    \textbf{Ours (A+E) (B3)} & \checkmark & \checkmark & \textbf{0.892} & 0.857 & \textbf{0.846} & \textbf{0.925} & \textbf{0.031} & 0.932 & 0.915 & 0.913 & \textbf{0.951} & \textbf{0.018} & \textbf{0.897} & \textbf{0.870} & \textbf{0.855} & \textbf{0.937} & \textbf{0.029} \\
    \bottomrule
  \end{tabular}
  }
\end{table*}

\subsubsection{Effectiveness of Uncertainty-Guided Dual-Stream Decoding.}

Table~\ref{tab:ablation_ugd} evaluates UGDD under both standard and degraded settings using $F_{\beta}^{w}$. Removing UGDD decreases performance from 0.913 to 0.908 on VT1000 and from 0.735 to 0.728 on VT-IMAG, while No-Unc. (dual-stream decoding without uncertainty guidance) also drops to 0.903 under degradations. Table~\ref{tab:response_u} further shows that $U_l^m$ responds to corrupted modalities, e.g., $U_l^r$ rises to 0.57 under RGB low-light, while $U_l^t$ rises to 0.55 under thermal contrast. These results verify the effectiveness of uncertainty-guided correction.

\begin{table}[t]
  \centering
  \caption{Ablation study of UGDD using $F_\beta^w$. ``Deg.'' averages five RGB/T degradation settings on VT1000.}
  \label{tab:ablation_ugd}
  \small
  \setlength{\tabcolsep}{11.5pt}
  \begin{tabular}{l c c c c}
    \toprule
    Method & VT1000 & VT5000 & VT-IMAG & Deg. \\
    \midrule
    w/o UGDD & 0.908 & 0.849 & 0.728 & 0.897 \\
    No-Unc.  & 0.911 & 0.852 & 0.731 & 0.903 \\
    \textbf{Full (Ours)} & \textbf{0.913} & \textbf{0.855} & \textbf{0.735} & \textbf{0.907} \\
    \bottomrule
  \end{tabular}
\end{table}

\begin{table}[!t]
  \centering
  \caption{Mean spatial $U_l^m$ on VT1000 under synthetic corruptions. The gate rises only for the degraded
modality.}
  \label{tab:response_u}
  \small
  \setlength{\tabcolsep}{7pt}
  \begin{tabular}{lcccccc}
    \toprule
    Cond. & Clean & Low-light & Noise & Blur & Contrast & Hot-block \\
    \midrule
    $U_l^r$ & 0.32 & \textbf{0.57} & \textbf{0.48} & \textbf{0.51} & 0.33 & 0.32 \\
    $U_l^t$ & 0.30 & 0.31 & 0.30 & 0.31 & \textbf{0.55} & \textbf{0.53} \\
    \bottomrule
  \end{tabular}
  \vskip -0.05in
\end{table}

\subsubsection{Effectiveness of Reliability-Conditioned Backbone Representation.}
Table~\ref{tab:abrcbr} reports the ablation results of the Reliability-Conditioned Backbone Representation (RCBR). 
Introducing modality-specific adapters (B1) and the shared expert bank (B2) consistently improves performance over the frozen backbone (B0). 
The full model (B3) achieves the best results, particularly on VT5000, improving $F_\beta^w$ from $0.851$ to $0.855$ and reducing $M$ to $0.029$, demonstrating the effectiveness of routing-aware residual modulation for handling modality reliability variations.

\subsubsection{Impact of the Number of Experts.}
Table~\ref{tab:ablation_expert} analyzes the effect of the number of experts $K$. 
The best trade-off between accuracy and robustness is achieved at $K=3$, which yields the highest $S_m$ (0.897) and $F_\beta^w$ (0.855) and the lowest $\mathcal{M}$ (0.029) on VT5000. 
Smaller $K$ limits transformation diversity, while larger $K$ leads to performance saturation and slight degradation, likely due to redundancy and increased optimization difficulty. 
Therefore, we set $K=3$ as the default configuration.

\begin{table*}[t]
  \centering
  \caption{Ablation on the number of experts $K$. Best results are highlighted in \textbf{bold}.}
  \resizebox{\textwidth}{!}{
  \begin{tabular}{c ccccc ccccc ccccc}
    \toprule
    \multirow{2}{*}{$K$} & \multicolumn{5}{c}{VT821} & \multicolumn{5}{c}{VT1000} & \multicolumn{5}{c}{VT5000} \\
    \cmidrule(lr){2-6} \cmidrule(lr){7-11} \cmidrule(lr){12-16}
    & $S_m\uparrow$ & $F_\beta\uparrow$ & $F_\beta^w\uparrow$ & $E_m\uparrow$ & $\mathcal{M}\downarrow$ & $S_m\uparrow$ & $F_\beta\uparrow$ & $F_\beta^w\uparrow$ & $E_m\uparrow$ & $\mathcal{M}\downarrow$ & $S_m\uparrow$ & $F_\beta\uparrow$ & $F_\beta^w\uparrow$ & $E_m\uparrow$ & $\mathcal{M}\downarrow$ \\
    \midrule
    1 & 0.890 & 0.856 & 0.844 & 0.925 & 0.032 & \textbf{0.933} & 0.914 & 0.912 & 0.950 & 0.019 & 0.896 & 0.868 & 0.853 & 0.936 & 0.030 \\
    2 & 0.891 & 0.856 & 0.844 & 0.924 & \textbf{0.031} & 0.932 & 0.913 & 0.911 & 0.949 & 0.019 & 0.896 & 0.858 & 0.854 & 0.933 & 0.030 \\
    3 & \textbf{0.892} & 0.857 & \textbf{0.846} & \textbf{0.925} & \textbf{0.031} & \textbf{0.932} & \textbf{0.915} & \textbf{0.913} & 0.951 & \textbf{0.018} & \textbf{0.897} & 0.870 & \textbf{0.855} & \textbf{0.937} & \textbf{0.029} \\
    4 & 0.890 & \textbf{0.859} & 0.844 & 0.923 & 0.032 & 0.930 & 0.912 & 0.911 & 0.950 & 0.019 & 0.895 & \textbf{0.871} & 0.852 & 0.935 & 0.030 \\
    5 & 0.889 & 0.855 & 0.842 & 0.921 & 0.032 & 0.929 & 0.909 & 0.909 & \textbf{0.952} & 0.020 & 0.894 & 0.868 & 0.850 & 0.933 & 0.031 \\
    \bottomrule
    \label{tab:ablation_expert}
  \end{tabular}
  }
\end{table*}

\subsubsection{Effectiveness of Pixel-wise Modality Competition.}
We compare the proposed Pixel-wise Modality Competition (PMC) with the MDAM module used in ConTriNet~\cite{tang2024divide}. As shown in Table~\ref{tab:ablation_pfd_internal}, replacing MDAM with PMC consistently improves performance across all benchmarks. On VT5000, PMC increases $F_\beta$ from $0.865$ to $0.870$ and $F_\beta^w$ from $0.850$ to $0.855$, with similar gains observed on VT821 ($E_m$: $0.919 \rightarrow 0.925$). These results suggest that pixel-wise competition better models spatially varying reliability, enabling adaptive modality selection and more accurate saliency predictions.


\begin{table*}[!htbp]
  \centering
    \caption{Ablation study of the proposed PMC compared with the baseline MDAM~\cite{tang2024divide}. Best results are highlighted in \textbf{bold}.}
  \resizebox{\textwidth}{!}{
  \begin{tabular}{lc ccccc ccccc ccccc}
    \toprule
    \multirow{2}{*}{Variant} & \multicolumn{5}{c}{VT821} & \multicolumn{5}{c}{VT1000} & \multicolumn{5}{c}{VT5000} \\
    \cmidrule(lr){3-7} \cmidrule(lr){8-12} \cmidrule(lr){13-17}
    & & $S_m \uparrow$ & $F_\beta \uparrow$ & $F_\beta^w \uparrow$ & $E_m \uparrow$ & $M \downarrow$ 
    & $S_m \uparrow$ & $F_\beta \uparrow$ & $F_\beta^w \uparrow$ & $E_m \uparrow$ & $M \downarrow$ 
    & $S_m \uparrow$ & $F_\beta \uparrow$ & $F_\beta^w \uparrow$ & $E_m \uparrow$ & $M \downarrow$ \\
    \midrule
    MDAM~\cite{tang2024divide} & \xmark & 0.890 & 0.853 & 0.840 & 0.919 & \textbf{0.030} & \textbf{0.932} & 0.913 & 0.911 & \textbf{0.951} & \textbf{0.018} & 0.896 & 0.865 & 0.850 & 0.933 & \textbf{0.029} \\
    PMC(Ours)   & \cmark & \textbf{0.892} & \textbf{0.857} & \textbf{0.846} & \textbf{0.925} & 0.031 & \textbf{0.932} & \textbf{0.915} & \textbf{0.913} & \textbf{0.951} & \textbf{0.018} & \textbf{0.897} & \textbf{0.870} & \textbf{0.855} & \textbf{0.937} & \textbf{0.029} \\
    \bottomrule
    \label{tab:ablation_pfd_internal}
  \end{tabular}
}
\end{table*}

\section{Conclusion}

In this work, we present RA-SOD, a reliability-aware RGB-T salient object detection framework designed to address modality degradation in real-world environments. 
Unlike existing approaches that mainly focus on fusion strategies, RA-SOD treats reliability as a fundamental principle and models it across representation learning, decoding refinement, and fusion. 
Specifically, a reliability-conditioned backbone adaptively compensates degraded features while preserving structural priors, an uncertainty-guided dual-stream decoder prevents unreliable cues from being propagated, and a pixel-wise modality competition strategy enables spatially adaptive cross-modal integration. 
Extensive experiments on multiple benchmarks demonstrate that RA-SOD achieves state-of-the-art performance and strong robustness under severe degradation conditions, highlighting the importance of reliability modeling for robust cross-modal perception.

\par\vfill\par


%
%
\bibliographystyle{splncs04}
\bibliography{main}

@String(CVPR  = {IEEE Conf. Comput. Vis. Pattern Recog.})

@String(ICCV  = {Int. Conf. Comput. Vis.})

@String(AAAI  = {AAAI})

@String(IJCAI = {IJCAI})

@String(ICME  = {Int. Conf. Multimedia and Expo})

@String(CVPR  = {CVPR})

@String(ICCV  = {ICCV})

@String(ICME  =	{ICME})

@inproceedings{mm/Xu000S023,
  author       = {Rui Xu and
                  Yong Luo and
                  Han Hu and
                  Bo Du and
                  Jialie Shen and
                  Yonggang Wen},
  title        = {Rethinking the Localization in Weakly Supervised Object Localization},
  booktitle    = {{ACM MM}},
  pages        = {5484--5494},
  year         = {2023}
}

@article{han2022trusted,
  title={Trusted multi-view classification with dynamic evidential fusion},
  author={Han, Zongbo and Zhang, Changqing and Fu, Huazhu and Zhou, Joey Tianyi},
  journal={IEEE transactions on pattern analysis and machine intelligence},
  volume={45},
  number={2},
  pages={2551--2566},
  year={2022},
  publisher={IEEE}
}

@article{wang2024rgb,
  title={RGB-T object detection with failure scenarios},
  author={Wang, Qingwang and Sun, Yuxuan and Chi, Yongke and Shen, Tao},
  journal={IEEE Journal of Selected Topics in Applied Earth Observations and Remote Sensing},
  volume={18},
  pages={3000--3010},
  year={2024},
  publisher={IEEE}
}

@article{tian2025learning,
  title={Learning a robust RGB-Thermal detector for extreme modality imbalance},
  author={Tian, Chao and Yang, Chao and Zhu, Guoqing and Wang, Qiang and He, Zhenyu},
  journal={Pattern Recognition Letters},
  volume={196},
  pages={1--8},
  year={2025},
  publisher={Elsevier}
}

@article{hu2025missingness,
  title={Missingness-aware prompting for modality-missing RGBT tracking},
  author={Hu, Guyue and Wang, Zhanghuan and Li, Chenglong and Yuan, Duzhi and He, Bin and Tang, Jin},
  journal={Journal of King Saud University Computer and Information Sciences},
  volume={37},
  number={6},
  pages={128},
  year={2025},
  publisher={Springer}
}

@article{wang2025confidence,
  title={Confidence-driven unimodal interference removal for enhanced multimodal object detection},
  author={Wang, Yu and Wei, Shikui and Xu, Sen and Qin, Ying and Zhao, Yao},
  journal={IEEE Transactions on Circuits and Systems for Video Technology},
  year={2025},
  publisher={IEEE}
}

@inproceedings{hao2024cola,
  title={Cola: Conditional dropout and language-driven robust dual-modal salient object detection},
  author={Hao, Shuang and Zhong, Chunlin and Tang, He},
  booktitle={European Conference on Computer Vision},
  pages={354--371},
  year={2024},
  organization={Springer}
}

@article{huo2022real,
  title={Real-time one-stream semantic-guided refinement network for RGB-thermal salient object detection},
  author={Huo, Fushuo and Zhu, Xuegui and Zhang, Qian and Liu, Ziming and Yu, Wenchao},
  journal={IEEE Transactions on Instrumentation and Measurement},
  volume={71},
  pages={1--12},
  year={2022},
  publisher={IEEE}
}

@article{huo2021efficient,
  title={Efficient context-guided stacked refinement network for RGB-T salient object detection},
  author={Huo, Fushuo and Zhu, Xuegui and Zhang, Lei and Liu, Qifeng and Shu, Yu},
  journal={IEEE Transactions on Circuits and Systems for Video Technology},
  volume={32},
  number={5},
  pages={3111--3124},
  year={2021},
  publisher={IEEE}
}

@article{wang2023thermal,
  title={Thermal images-aware guided early fusion network for cross-illumination RGB-T salient object detection},
  author={Wang, Han and Song, Kechen and Huang, Liming and Wen, Hongwei and Yan, Yunhui},
  journal={Engineering Applications of Artificial Intelligence},
  volume={118},
  pages={105640},
  year={2023},
  publisher={Elsevier}
}

@ARTICLE{10032588,
  author={Xie, Zhengxuan and Shao, Feng and Chen, Gang and Chen, Hangwei and Jiang, Qiuping and Meng, Xiangchao and Ho, Yo-Sung},
  journal={IEEE Transactions on Circuits and Systems for Video Technology}, 
  title={Cross-Modality Double Bidirectional Interaction and Fusion Network for RGB-T Salient Object Detection}, 
  year={2023},
  volume={33},
  number={8},
  pages={4149-4163},
  doi={10.1109/TCSVT.2023.3241196}}

@INPROCEEDINGS{11093604,
  author={He, Jiahao and Fu, Keren and Liu, Xiaohong and Zhao, Qijun},
  booktitle={2025 IEEE/CVF Conference on Computer Vision and Pattern Recognition (CVPR)}, 
  title={Samba: A Unified Mamba-Based Framework for General Salient Object Detection}, 
  year={2025},
  volume={},
  number={},
  pages={25314-25324},
  doi={10.1109/CVPR52734.2025.02357}}

@ARTICLE{10006743,
  author={Ma, Mingcan and Xia, Changqun and Xie, Chenxi and Chen, Xiaowu and Li, Jia},
  journal={IEEE Transactions on Image Processing}, 
  title={Boosting Broader Receptive Fields for Salient Object Detection}, 
  year={2023},
  volume={32},
  number={},
  pages={1026-1038},
  doi={10.1109/TIP.2022.3232209}}

@ARTICLE{10504918,
  author={Lv, Chengtao and Zhou, Xiaofei and Wan, Bin and Wang, Shuai and Sun, Yaoqi and Zhang, Jiyong and Yan, Chenggang},
  journal={IEEE Transactions on Consumer Electronics}, 
  title={Transformer-Based Cross-Modal Integration Network for RGB-T Salient Object Detection}, 
  year={2024},
  volume={70},
  number={2},
  pages={4741-4755},
  doi={10.1109/TCE.2024.3390841}}

@ARTICLE{10587282,
  author={Wang, Yue and Zhang, Lu and Zhang, Pingping and Zhuge, Yunzhi and Wu, Junfeng and Yu, Hong and Lu, Huchuan},
  journal={IEEE Transactions on Circuits and Systems for Video Technology}, 
  title={Learning Local-Global Representation for Scribble-Based RGB-D Salient Object Detection via Transformer}, 
  year={2024},
  volume={34},
  number={11},
  pages={11592-11604},
  doi={10.1109/TCSVT.2024.3424651}}

@ARTICLE{11131311,
  author={Wang, Huizhi and Guo, Hui and Chai, Xiongli and Mu, Baoyang and Shao, Feng},
  journal={IEEE Transactions on Instrumentation and Measurement}, 
  title={Cognition-Inspired Dynamic Feature Integration Network for RGB-D and RGB-T Salient Object Detection}, 
  year={2025},
  volume={74},
  number={},
  pages={1-16},
  doi={10.1109/TIM.2025.3600718}}

@ARTICLE{10127616,
  author={Zhou, Wujie and Sun, Fan and Jiang, Qiuping and Cong, Runmin and Hwang, Jenq-Neng},
  journal={IEEE Transactions on Image Processing}, 
  title={WaveNet: Wavelet Network With Knowledge Distillation for RGB-T Salient Object Detection}, 
  year={2023},
  volume={32},
  number={},
  pages={3027-3039},
  doi={10.1109/TIP.2023.3275538}}

@article{wan2024mffnet,
  title={Mffnet: Multi-modal feature fusion network for v-d-t salient object detection},
  author={Wan, Bo and Zhou, Xueping and Sun, Yanan and Wang, Tian and Lv, Chen and Wang, Shuang and Yin, Haijian and Yan, Chenggang},
  journal={IEEE Transactions on Multimedia},
  volume={26},
  pages={2069--2081},
  year={2024},
  publisher={IEEE}
}

@inproceedings{luo2024vscode,
  title={VSCode: General Visual Salient and Camouflaged Object Detection with 2D Prompt Learning},
  author={Luo, Ziyang and Liu, Nian and Zhao, Wangbo and Yang, Xuguang and Zhang, Dingwen and Fan, Deng-Ping and Khan, Fahad and Han, Junwei},
  booktitle={Proceedings of the IEEE/CVF Conference on Computer Vision and Pattern Recognition (CVPR)},
  pages={17169--17180},
  year={2024}
}

@ARTICLE{10003255,
  author={Song, Kechen and Huang, Liming and Gong, Aojun and Yan, Yunhui},
  journal={IEEE Transactions on Circuits and Systems for Video Technology}, 
  title={Multiple Graph Affinity Interactive Network and a Variable Illumination Dataset for RGBT Image Salient Object Detection}, 
  year={2023},
  volume={33},
  number={7},
  pages={3104-3118},
  doi={10.1109/TCSVT.2022.3233131}}

@ARTICLE{8603756,
  author={Chen, Hao and Li, Youfu},
  journal={IEEE Transactions on Image Processing}, 
  title={Three-Stream Attention-Aware Network for RGB-D Salient Object Detection}, 
  year={2019},
  volume={28},
  number={6},
  pages={2825-2835},
  doi={10.1109/TIP.2019.2891104}}

@inproceedings{yin2025dformerv2,
  title={DFormerv2: Geometry Self-Attention for RGBD Semantic Segmentation},
  author={Yin, Bo-Wen and Cao, Jiao-Long and Cheng, Ming-Ming and Hou, Qibin},
  booktitle={Proceedings of the Computer Vision and Pattern Recognition Conference},
  pages={19345--19355},
  year={2025}
}

@INPROCEEDINGS{7780626,
  author={Feng, David and Barnes, Nick and You, Shaodi and McCarthy, Chris},
  booktitle={2016 IEEE Conference on Computer Vision and Pattern Recognition (CVPR)}, 
  title={Local Background Enclosure for RGB-D Salient Object Detection}, 
  year={2016},
  volume={},
  number={},
  pages={2343-2350},
  doi={10.1109/CVPR.2016.257}}

@INPROCEEDINGS{8954074,
  author={Zhao, Ting and Wu, Xiangqian},
  booktitle={2019 IEEE/CVF Conference on Computer Vision and Pattern Recognition (CVPR)}, 
  title={Pyramid Feature Attention Network for Saliency Detection}, 
  year={2019},
  volume={},
  number={},
  pages={3080-3089},
  doi={10.1109/CVPR.2019.00320}}

@inproceedings{zhao2019egnet,
  title={EGNet: Edge guidance network for salient object detection},
  author={Zhao, Jia-Xing and Liu, Jiang-Jiang and Fan, Deng-Ping and Cao, Yang and Yang, Jufeng and Cheng, Ming-Ming},
  booktitle={Proceedings of the IEEE/CVF international conference on computer vision},
  pages={8779--8788},
  year={2019}
}

@INPROCEEDINGS{8578428,
  author={Wang, Tiantian and Zhang, Lihe and Wang, Shuo and Lu, Huchuan and Yang, Gang and Ruan, Xiang and Borji, Ali},
  booktitle={2018 IEEE/CVF Conference on Computer Vision and Pattern Recognition}, 
  title={Detect Globally, Refine Locally: A Novel Approach to Saliency Detection}, 
  year={2018},
  volume={},
  number={},
  pages={3127-3135},
  doi={10.1109/CVPR.2018.00330}}

@INPROCEEDINGS{8237695,
  author={Wang, Tiantian and Borji, Ali and Zhang, Lihe and Zhang, Pingping and Lu, Huchuan},
  booktitle={2017 IEEE International Conference on Computer Vision (ICCV)}, 
  title={A Stagewise Refinement Model for Detecting Salient Objects in Images}, 
  year={2017},
  volume={},
  number={},
  pages={4039-4048},
  doi={10.1109/ICCV.2017.433}}

@ARTICLE{9076883,
  author={Liu, Nian and Han, Junwei and Yang, Ming-Hsuan},
  journal={IEEE Transactions on Image Processing}, 
  title={PiCANet: Pixel-Wise Contextual Attention Learning for Accurate Saliency Detection}, 
  year={2020},
  volume={29},
  number={},
  pages={6438-6451},
  doi={10.1109/TIP.2020.2988568}}

@inproceedings{zhang2017learning,
  title={Learning uncertain convolutional features for accurate saliency detection},
  author={Zhang, Pingping and Wang, Dong and Lu, Huchuan and Wang, Hongyu and Yin, Baocai},
  booktitle={Proceedings of the IEEE International Conference on computer vision},
  pages={212--221},
  year={2017}
}

@inproceedings{zhang2017amulet,
  title={Amulet: Aggregating multi-level convolutional features for salient object detection},
  author={Zhang, Pingping and Wang, Dong and Lu, Huchuan and Wang, Hongyu and Ruan, Xiang},
  booktitle={Proceedings of the IEEE international conference on computer vision},
  pages={202--211},
  year={2017}
}

@ARTICLE{7488288,
  author={Li, Xi and Zhao, Liming and Wei, Lina and Yang, Ming-Hsuan and Wu, Fei and Zhuang, Yueting and Ling, Haibin and Wang, Jingdong},
  journal={IEEE Transactions on Image Processing}, 
  title={DeepSaliency: Multi-Task Deep Neural Network Model for Salient Object Detection}, 
  year={2016},
  volume={25},
  number={8},
  pages={3919-3930},
  doi={10.1109/TIP.2016.2579306}}

@INPROCEEDINGS{6619110,
  author={Jiang, Zhuolin and Davis, Larry S.},
  booktitle={2013 IEEE Conference on Computer Vision and Pattern Recognition}, 
  title={Submodular Salient Region Detection}, 
  year={2013},
  volume={},
  number={},
  pages={2043-2050},
  doi={10.1109/CVPR.2013.266}}

@ARTICLE{7307162,
  author={Kim, Jiwhan and Han, Dongyoon and Tai, Yu-Wing and Kim, Junmo},
  journal={IEEE Transactions on Image Processing}, 
  title={Salient Region Detection via High-Dimensional Color Transform and Local Spatial Support}, 
  year={2016},
  volume={25},
  number={1},
  pages={9-23},
  doi={10.1109/TIP.2015.2495122}}

@ARTICLE{6871397,
  author={Cheng, Ming-Ming and Mitra, Niloy J. and Huang, Xiaolei and Torr, Philip H. S. and Hu, Shi-Min},
  journal={IEEE Transactions on Pattern Analysis and Machine Intelligence}, 
  title={Global Contrast Based Salient Region Detection}, 
  year={2015},
  volume={37},
  number={3},
  pages={569-582},
  doi={10.1109/TPAMI.2014.2345401}}

@article{cong2022does,
  title={Does thermal really always matter for RGB-T salient object detection?},
  author={Cong, Runmin and Zhang, Kepu and Zhang, Chen and Zheng, Feng and Zhao, Yao and Huang, Qingming and Kwong, Sam},
  journal={IEEE Transactions on Multimedia},
  volume={25},
  pages={6971--6982},
  year={2022},
  publisher={IEEE}
}

@inproceedings{liu2023scribble,
  title={Scribble-supervised RGB-T salient object detection},
  author={Liu, Zhengyi and Huang, Xiaoshen and Zhang, Guanghui and Fang, Xianyong and Wang, Linbo and Tang, Bin},
  booktitle={2023 IEEE International Conference on Multimedia and Expo (ICME)},
  pages={2369--2374},
  year={2023},
  organization={IEEE}
}

@ARTICLE{9161021,
  author={Zhang, Qiang and Xiao, Tonglin and Huang, Nianchang and Zhang, Dingwen and Han, Jungong},
  journal={IEEE Transactions on Circuits and Systems for Video Technology}, 
  title={Revisiting Feature Fusion for RGB-T Salient Object Detection}, 
  year={2021},
  volume={31},
  number={5},
  pages={1804-1818},
  doi={10.1109/TCSVT.2020.3014663}}

@ARTICLE{8935533,
  author={Zhang, Qiang and Huang, Nianchang and Yao, Lin and Zhang, Dingwen and Shan, Caifeng and Han, Jungong},
  journal={IEEE Transactions on Image Processing}, 
  title={RGB-T Salient Object Detection via Fusing Multi-Level CNN Features}, 
  year={2020},
  volume={29},
  number={},
  pages={3321-3335},
  doi={10.1109/TIP.2019.2959253}}

@inproceedings{fu2020jl,
  title={JL-DCF: Joint learning and densely-cooperative fusion framework for RGB-D salient object detection},
  author={Fu, Keren and Fan, Deng-Ping and Ji, Ge-Peng and Zhao, Qijun},
  booktitle={Proceedings of the IEEE/CVF Conference on Computer Vision and Pattern Recognition (CVPR)},
  pages={3052--3062},
  year={2020}
}

@article{wang2022learning,
  title={Learning discriminative cross-modality features for RGB-D saliency detection},
  author={Wang, Feng and Pan, Jinjin and Xu, Shihan and Tang, Jinhui},
  journal={IEEE Transactions on Image Processing},
  volume={31},
  pages={1285--1297},
  year={2022},
  publisher={IEEE}
}

@article{cong2022cir,
  title={CIR-Net: Cross-modality interaction and refinement for RGB-D salient object detection},
  author={Cong, Runmin and Lin, Qinwei and Zhang, Chen and Li, Chongyi and Zhao, Yao and Kwong, Sam and Hoi, Steven CH},
  journal={IEEE Transactions on Image Processing},
  volume={31},
  pages={6800--6815},
  year={2022},
  publisher={IEEE}
}

@inproceedings{wu2022robust,
  title={Robust RGB-D fusion for saliency detection},
  author={Wu, Zhe and Gobichettipalayam, Saihui and Tamadazte, Brahim and Allibert, Guillaume and Paudel, Danda Pani and Demonceaux, C{\'e}dric},
  booktitle={International Conference on 3D Vision (3DV)},
  pages={403--413},
  year={2022},
  organization={IEEE}
}

@inproceedings{wu2023source,
  title={Source-free depth for object pop-out},
  author={Wu, Zongwei and Allibert, Guillaume and Meriaudeau, Fabrice and Demonceaux, Cedric},
  booktitle={Proceedings of the IEEE/CVF International Conference on Computer Vision (ICCV)},
  pages={1032--1042},
  year={2023}
}

@article{wu2023hidanet,
  title={HiDAnet: RGB-D salient object detection via hierarchical depth awareness},
  author={Wu, Zongwei and Allibert, Guillaume and M{\'e}riaudeau, Fabrice and Ma, Chao and Demonceaux, C{\'e}dric},
  journal={IEEE Transactions on Image Processing},
  volume={32},
  pages={2160--2173},
  year={2023},
  publisher={IEEE}
}

@inproceedings{wang2018rgb,
  title={RGB-T saliency detection benchmark: Dataset, baselines, analysis and a novel approach},
  author={Wang, Guoxia and Li, Cong and Ma, Yunyao and Zheng, Aihua and Tang, Jinhui and Luo, Bin},
  booktitle={China Conference on Image and Graphics Technologies and Applications},
  pages={359--369},
  year={2018},
  organization={Springer}
}

@inproceedings{tu2019m3s,
  title={M3S-NIR: Multi-modal multi-scale noise-insensitive ranking for RGB-T saliency detection},
  author={Tu, Zhengzheng and Xia, Tian and Li, Chang and Lu, Yan and Tang, Jinhui},
  booktitle={IEEE Conference on Multimedia Information Processing and Retrieval (MIPR)},
  pages={141--146},
  year={2019}
}

@article{tu2020rgb,
  title={RGB-T image saliency detection via collaborative graph learning},
  author={Tu, Zhengzheng and Xia, Tian and Li, Chang and Wang, Xiao and Ma, Yan and Tang, Jinhui},
  journal={IEEE Transactions on Multimedia},
  volume={22},
  number={1},
  pages={160--173},
  year={2020},
  publisher={IEEE}
}

@article{tu2022rgbt,
  title={RGBT salient object detection: A large-scale dataset and benchmark},
  author={Tu, Zhengzheng and Ma, Yunyao and Li, Zechao and Li, Chang and Xu, Jiasheng and Liu, Yong},
  journal={IEEE Transactions on Multimedia},
  year={2022},
  note={Also arXiv:2007.03262}
}

@article{tu2021multi,
  title={Multi-interactive dual-decoder for RGB-thermal salient object detection},
  author={Tu, Zhengzheng and Li, Zechao and Li, Chang and Lang, Yuan and Tang, Jinhui},
  journal={IEEE Transactions on Image Processing},
  volume={30},
  pages={5678--5691},
  year={2021},
  publisher={IEEE}
}

@article{wang2022cgfnet,
  title={CGFNet: Cross-guided fusion network for RGB-T salient object detection},
  author={Wang, Jinyu and Song, Keren and Bao, Yuxuan and Huang, Liming and Yan, Yunhui},
  journal={IEEE Transactions on Circuits and Systems for Video Technology},
  volume={32},
  number={5},
  pages={2949--2961},
  year={2022},
  publisher={IEEE}
}

@article{gao2022unified,
  title={Unified information fusion network for multi-modal RGB-D and RGB-T salient object detection},
  author={Gao, Wei and Liao, Guibiao and Ma, Siqi and Li, Ge and Liang, Yijun and Lin, Weisi},
  journal={IEEE Transactions on Circuits and Systems for Video Technology},
  volume={32},
  number={4},
  pages={2091--2106},
  year={2022},
  publisher={IEEE}
}

@article{liang2022multi,
  title={Multi-modal interactive attention and dual progressive decoding network for RGB-D/T salient object detection},
  author={Liang, Yijun and Qin, Ge and Sun, Ming and Qin, Jing and Yan, Jun and Zhang, Zhicheng},
  journal={Neurocomputing},
  volume={490},
  pages={132--145},
  year={2022},
  publisher={Elsevier}
}

@article{zhou2022ecffnet,
  title={ECFFNet: Effective and consistent feature fusion network for RGB-T salient object detection},
  author={Zhou, Wujie and Guo, Qing and Lei, Jiansheng and Yu, Ling and Hwang, Jenq-Neng},
  journal={IEEE Transactions on Circuits and Systems for Video Technology},
  volume={32},
  number={3},
  pages={1224--1235},
  year={2022},
  publisher={IEEE}
}

@article{zhou2023lsnet,
  title={LSNet: Lightweight spatial boosting network for detecting salient objects in RGB-thermal images},
  author={Zhou, Wujie and Zhu, Yifan and Lei, Jiansheng and Yang, Ren and Yu, Ling},
  journal={IEEE Transactions on Image Processing},
  volume={32},
  pages={1329--1340},
  year={2023},
  publisher={IEEE}
}

@article{pang2023caver,
  title={CAVER: Cross-modal view-mixed transformer for bi-modal salient object detection},
  author={Pang, Yanwei and Zhao, Xiaoqi and Zhang, Lihe and Lu, Huchuan},
  journal={IEEE Transactions on Image Processing},
  volume={32},
  pages={892--904},
  year={2023},
  publisher={IEEE}
}

@article{tang2024divide,
  author={Tang, Hao and Li, Zechao and Zhang, Dong and He, Shengfeng and Tang, Jinhui},
  journal={IEEE Transactions on Pattern Analysis and Machine Intelligence}, 
  title={Divide-and-Conquer: Confluent Triple-Flow Network for RGB-T Salient Object Detection}, 
  year={2025},
  volume={47},
  number={3},
  pages={1958-1974},
  doi={10.1109/TPAMI.2024.3511621}}

@article{wang2024learning,
  title={Learning adaptive fusion bank for multi-modal salient object detection},
  author={Wang, Kunpeng and Tu, Zhengzheng and Li, Chenglong and Zhang, Cheng and Luo, Bin},
  journal={IEEE Transactions on Circuits and Systems for Video Technology},
  volume={34},
  number={8},
  pages={7344--7358},
  year={2024},
  publisher={IEEE}
}

@inproceedings{fan2017structure,
  title={Structure-measure: A New Way to Evaluate Foreground Maps},
  author={Fan, Deng-Ping and Cheng, Ming-Ming and Liu, Yun and Li, Tao and Borji, Ali},
  booktitle={Proceedings of the IEEE International Conference on Computer Vision (ICCV)},
  pages={4548--4557},
  year={2017}
}

@inproceedings{achanta2009frequency,
  title={Frequency-tuned Salient Region Detection},
  author={Achanta, Radhakrishna and Hemami, Sheila and Estrada, Francisco and Susstrunk, Sabine},
  booktitle={Proceedings of the IEEE Conference on Computer Vision and Pattern Recognition (CVPR)},
  pages={1597--1604},
  year={2009}
}

@inproceedings{margolin2014how,
  title={How to Evaluate Foreground Maps?},
  author={Margolin, Ran and Zelnik-Manor, Lihi and Tal, Ayellet},
  booktitle={Proceedings of the IEEE Conference on Computer Vision and Pattern Recognition (CVPR)},
  pages={248--255},
  year={2014}
}

@inproceedings{fan2018enhanced,
  title={Enhanced-alignment Measure for Binary Foreground Map Evaluation},
  author={Fan, Deng-Ping and Gong, Cheng and Cao, Yang and Ren, Bo and Cheng, Ming-Ming and Borji, Ali},
  booktitle={Proceedings of the International Joint Conference on Artificial Intelligence (IJCAI)},
  pages={698--704},
  year={2018}
}

@inproceedings{perazzi2012saliency,
  title={Saliency Filters: Contrast Based Filtering for Salient Region Detection},
  author={Perazzi, Federico and Kr{\"a}henb{\"u}hl, Philipp and Pritch, Yael and Hornung, Alexander},
  booktitle={Proceedings of the IEEE Conference on Computer Vision and Pattern Recognition (CVPR)},
  pages={733--740},
  year={2012}
}

@inproceedings{he2019bag,
  title={Bag of tricks for image classification with convolutional neural networks},
  author={He, Tong and Zhang, Zhi and Zhang, Hang and Zhang, Zhongyue and Xie, Junyuan and Li, Mu},
  booktitle={Proceedings of the IEEE/CVF conference on computer vision and pattern recognition},
  pages={558--567},
  year={2019}
}

@inproceedings{xiao2025smr,
  title={SMR-Net: Semantic-guided mutually reinforcing network for cross-modal image fusion and salient object detection},
  author={Xiao, Guobao and Liu, Xinyu and Lin, Zebin and Ming, Rui},
  booktitle={Proceedings of the AAAI Conference on Artificial Intelligence},
  volume={39},
  number={8},
  pages={8637--8645},
  year={2025}
}

@article{jacobs1991adaptive,
  title={Adaptive Mixtures of Local Experts},
  author={Jacobs, Robert A. and Jordan, Michael I. and Nowlan, Steven J. and Hinton, Geoffrey E.},
  journal={Neural Computation},
  volume={3},
  number={1},
  pages={79--87},
  year={1991}
}

@article{shazeer2017outrageously,
  title={Outrageously Large Neural Networks: The Sparsely-Gated Mixture-of-Experts Layer},
  author={Shazeer, Noam and Mirhoseini, Azalia and Maziarz, Krzysztof and Davis, Andy and Le, Quoc and Hinton, Geoffrey and Dean, Jeff},
  journal={arXiv preprint arXiv:1701.06538},
  year={2017}
}

@inproceedings{riquelme2021scaling,
  title={Scaling Vision with Sparse Mixture of Experts},
  author={Riquelme, Carlos and Puigcerver, Joan and Mustafa, Basil and Neumann, Maxim and Jenatton, Rodolphe and Pinto, Andr{\\'e} Susano and Keysers, Daniel and Houlsby, Neil},
  booktitle={Advances in Neural Information Processing Systems},
  volume={34},
  pages={8583--8595},
  year={2021}
}

@article{wang2024alignment,
  title={Alignment-free rgbt salient object detection: Semantics-guided asymmetric correlation network and a unified benchmark},
  author={Wang, Kunpeng and Lin, Danying and Li, Chenglong and Tu, Zhengzheng and Luo, Bin},
  journal={IEEE Transactions on Multimedia},
  volume={26},
  pages={10692--10707},
  year={2024},
  publisher={IEEE}
}

@inproceedings{wang2025alignment,
  title={Alignment-free rgb-t salient object detection: A large-scale dataset and progressive correlation network},
  author={Wang, Kunpeng and Chen, Keke and Li, Chenglong and Tu, Zhengzheng and Luo, Bin},
  booktitle={Proceedings of the AAAI conference on artificial intelligence},
  volume={39},
  number={7},
  pages={7780--7788},
  year={2025}
}

@article{tang2022hrtransnet,
  title={HRTransNet: HRFormer-driven two-modality salient object detection},
  author={Tang, Bin and Liu, Zhengyi and Tan, Yacheng and He, Qian},
  journal={IEEE Transactions on Circuits and Systems for Video Technology},
  volume={33},
  number={2},
  pages={728--742},
  year={2022},
  publisher={IEEE}
}

@inproceedings{wu2023object,
  title={Object segmentation by mining cross-modal semantics},
  author={Wu, Zongwei and Wang, Jingjing and Zhou, Zhuyun and An, Zhaochong and Jiang, Qiuping and Demonceaux, C{\'e}dric and Sun, Guolei and Timofte, Radu},
  booktitle={Proceedings of the 31st ACM International Conference on Multimedia},
  pages={3455--3464},
  year={2023}
}
\end{document}